%% file: main.tex
\documentclass[runningheads]{llncs}
\usepackage[T1]{fontenc}
\usepackage{graphicx,verbatim}
\usepackage{booktabs}
\usepackage{colortbl}
\usepackage{xcolor}
\usepackage{multirow}
\usepackage{amsmath}
\usepackage{amssymb}
\usepackage{microtype}
\begin{document}

\newcommand{\mydataset}{\textit{PedSkull-CT}}
\newcommand{\myit}{\textit{PSCT-Net}}

\newcommand{\rev}[1]{\textcolor{blue}{#1}}

\title{PSCT-Net: Geometry-Aware Pediatric Skull CT Reconstruction via Differentiable Back-Projection and Attention-Guided Refinement}
\titlerunning{PSCT-Net for Pediatric Skull CT Reconstruction}
%

\author{Dong Yeong Kim\inst{1,2} \and
Jaewon Choi\inst{2} \and
Youmin Shin\inst{1,2} \and
JunGyu Lee\inst{3}$^{\star}$ \and 
Myeongseop Kim\inst{2} \and
Jinwook Choi\inst{1,5} \and
Joo Whan Kim\inst{4,5}$^{\dagger}$ \and
Young-Gon Kim\inst{2,5}$^{\dagger}$}

\authorrunning{D.Y. Kim et al.}

\institute{
Interdisciplinary Program in Bioengineering, Seoul National University \and
Department of Transdisciplinary Medicine, Seoul National University Hospital \and
Department of Artificial Intelligence, Yonsei University \and
Division of Pediatric Neurosurgery, Seoul National University Children’s Hospital \and
Department of Medicine, Seoul National University College of Medicine \\
\email{steven6774@snu.ac.kr}
}

\maketitle              

{\let\thefootnote\relax\footnotetext{$^{\star}$ Work done prior to joining$^{3}$ \ $^{\dagger}$ Corresponding Author}}
\begin{abstract}

Computed Tomography (CT) is essential for diagnosing pediatric craniofacial abnormalities, yet poses radiation risks to developing anatomies. Reconstructing 3D CT from sparse bi-planar X-rays offers a low-dose alternative but is severely ill-posed. Existing methods employ geometry-agnostic feature lifting, naively projecting 2D features into 3D without explicit spatial modeling, causing depth ambiguity and degraded osseous boundaries. We present PSCT-Net, a geometry-aware framework with differentiable back-projection. Differentiable back-projection establishes a spatially faithful volumetric prior, alleviating depth ambiguity. An Attention-Guided Projection (AGP-3D) module then learns non-linear voxel-wise correspondences between 2D regions and 3D locations. A Bidirectional Mamba (BiM-3D) module captures long-range volumetric dependencies with linear complexity. We further curate a private institutional pediatric skull CT cohort, PedSkull-CT, comprising normal and pathological cases for internal evaluation, addressing the gap in adult-centric, trunk-focused datasets. Project page and code are available at \url{https://dydevelop.github.io/PSCT-Net/}.


\end{abstract}
\input{intro}
\input{methods}
\input{experiments}
\input{conclusion}

\subsubsection*{Acknowledgements.} 
This work was supported in part by the Seoul National University Hospital Research Fund (Grant No. 04-2024-0430).

\noindent\textbf{Disclosure of Interests.}
The authors have no competing interests to declare that are relevant to the content of this article.

%
%
%
\bibliographystyle{splncs04}
\bibliography{cas-refs}

\end{document}

%% file: intro.tex
\section{Introduction}

 Computed tomography (CT) stands as the gold standard for diagnosing craniofacial abnormalities, providing essential three-dimensional (3D) anatomical details for surgical planning~\cite{vannier1989craniosynostosis,mettler2008effective}. However, the substantial ionizing radiation dose associated with CT poses severe risks to pediatric populations, who exhibit heightened radiosensitivity and a longer lifespan for potential radiation-induced malignancies~\cite{brenner2007computed,goske2008image}. While bi-planar X-ray imaging offers a low-dose alternative~\cite{kim2025multi}, it fundamentally lacks the volumetric depth information required to assess complex skull deformities. Consequently, reconstructing high-fidelity 3D CT volumes from sparse 2D X-ray projections has emerged as a critical yet challenging objective in pediatric imaging.

This reconstruction task, however, remains a severely ill-posed inverse problem. Existing deep learning approaches~\cite{henzler2018single,shen2019patient,chen2023bx2s,ge2022x,ying2019x2ct,wang2023trct} typically rely on geometry-agnostic feature lifting, where 2D features are naively replicated or linearly projected into 3D space. Such implicit projection mechanisms fail to model the physical acquisition geometry, leading to spatial misalignment and the loss of fine-grained osseous structures (e.g., sutures and fontanelles) crucial for pediatric diagnosis. While recent diffusion-based models~\cite{liu2024diffux2ct,bai2024xctdiff,xie2025dvg} have improved textural realism, their iterative denoising processes impose prohibitive computational costs~\cite{song2020denoising}, rendering them impractical for time-sensitive clinical workflows.

\begin{figure}[t]
    \centering
    \includegraphics[width=\linewidth]{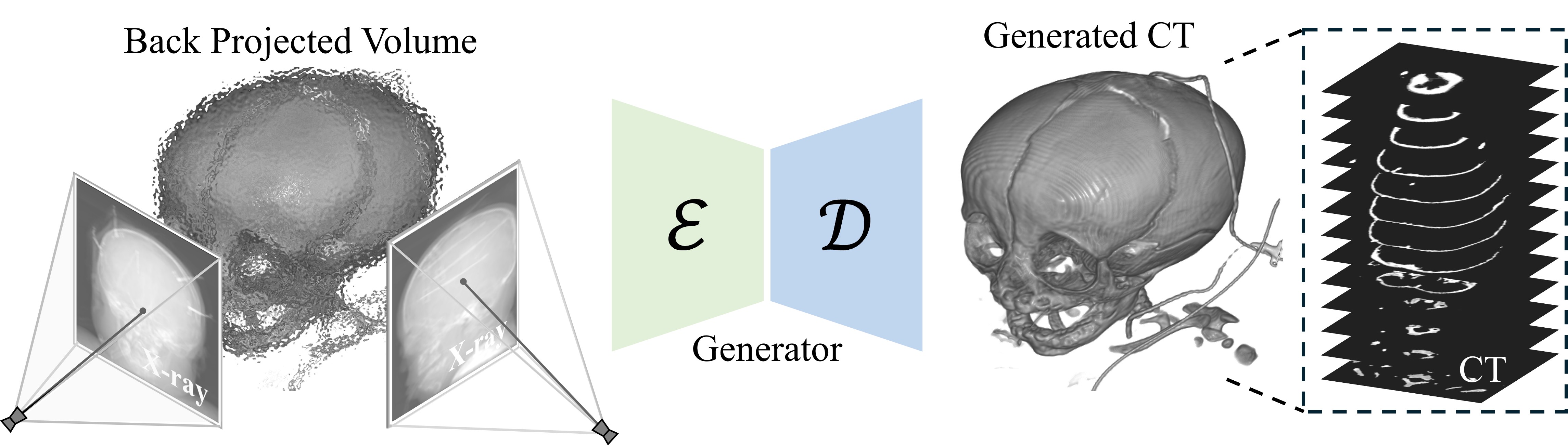}
    \caption{\textbf{Overview of the proposed approach.} Frontal and lateral X-rays are back-projected to form a coarse volumetric prior. This geometric prior is then refined by the generator using original X-rays to reconstruct a high-fidelity CT volume.}
    \label{fig:1}
\end{figure}

To address these limitations, we propose \textbf{PSCT-Net}, a novel framework that integrates explicit geometric priors with computationally efficient context modeling (Fig.~\ref{fig:1}). Unlike previous methods that treat 2D-to-3D lifting as a black-box transformation, our approach injects acquisition geometry directly into the network via \textbf{differentiable back-projection}, establishing a spatially faithful volumetric initialization that alleviates depth ambiguity from the outset. Building upon this geometric prior, an \textbf{Attention-Guided Projection (AGP-3D)} module learns non-linear voxel-wise correspondences between 2D image regions and 3D spatial locations, replacing naive linear projections. To capture holistic cranial geometry efficiently, we incorporate a \textbf{Bidirectional Mamba (BiM-3D)} module that models long-range volumetric dependencies with linear complexity, avoiding the quadratic cost of standard attention mechanisms.

Furthermore, we identify a critical two-fold domain gap that hinders progress in pediatric reconstruction. Existing public benchmarks~\cite{armato2004lung,deng2024ctspine1k,liu2021ctpelvic} focus exclusively on trunk anatomy (e.g., lung, spine, and pelvis), which possesses fundamentally different geometric topologies compared to the intricate structure of the cranium. Moreover, these datasets are adult-centric, failing to capture the unique physiological markers of the pediatric skull, such as unclosed fontanelles and thinner cortical bone~\cite{vannier1989craniosynostosis}. To study this setting, we curate \textbf{PedSkull-CT,} a private institutional pediatric skull CT cohort comprising both normal and pathological cases with paired simulated X-rays.


\noindent In summary, our contributions are threefold:

\begin{itemize}
    \item We propose PSCT-Net, the first framework to integrate differentiable back-projection with state-space modeling for X-ray to CT reconstruction. By explicitly encoding acquisition geometry, our method resolves depth ambiguity that causes existing approaches to hallucinate incorrect structures.
    \item We curate PedSkull-CT, a private institutional pediatric skull CT cohort with 982 scans covering both normal and pathological cases, and use it for internal validation.
    \item We establish state-of-the-art performance on the private PedSkull-CT cohort and achieve highly competitive results across three public benchmarks. Our method outperforms diffusion-based approaches in overall fidelity (PSNR) while maintaining the high inference efficiency inherent to single-step architectures, making it highly suitable for clinical deployment.
\end{itemize}

%% file: methods.tex
\section{Methodology}

\begin{figure*}[t]
    \centering
    \includegraphics[width=\textwidth]{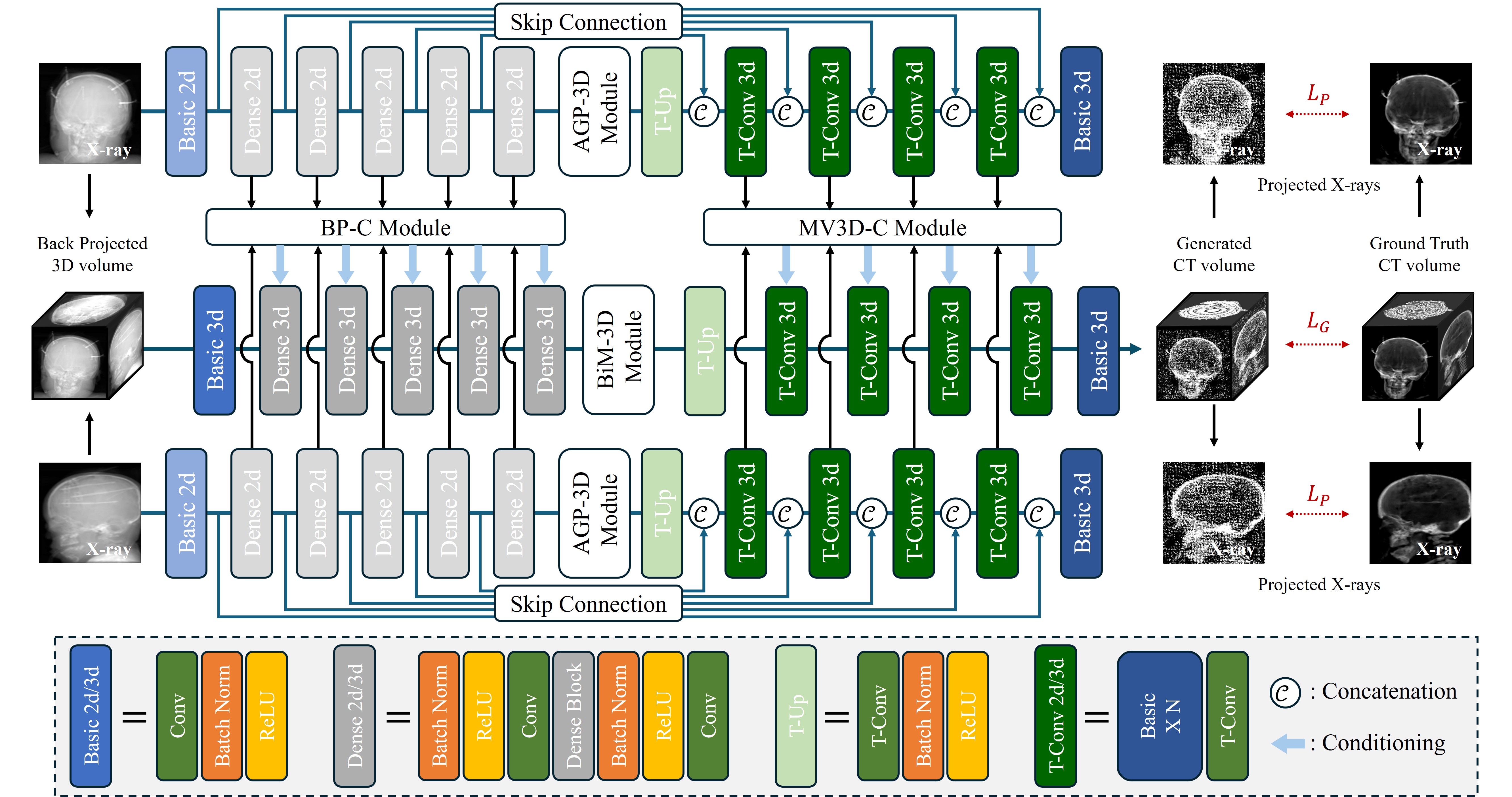}
    \caption{\textbf{Overview of PSCT-Net.} The framework initializes a coarse volumetric prior via differentiable back-projection. This prior is refined by an encoder-decoder explicitly conditioned by the BP-C and MV3D-C modules to enforce geometric consistency. The network is trained with adversarial, voxel-wise reconstruction($L_{G}$), and projection-consistency losses($L_{P}$); the adversarial branch is omitted from the diagram for clarity.}
    \label{fig:2}
\end{figure*}

We propose PSCT-Net for geometry-aware pediatric skull CT reconstruction. Our PSCT-Net integrates differentiable back-projection and attention-guided feature lifting into a standard 2D-to-3D cGAN framework. By further incorporating geometry-aware conditioning and a bidirectional state space bottleneck, we ensure both spatial consistency and robust global context modeling. (Fig. \ref{fig:2})

\subsection{Back-Projection Volumetric Initialization}

\begin{figure*}[t]
    \centering
    \includegraphics[width=\linewidth]{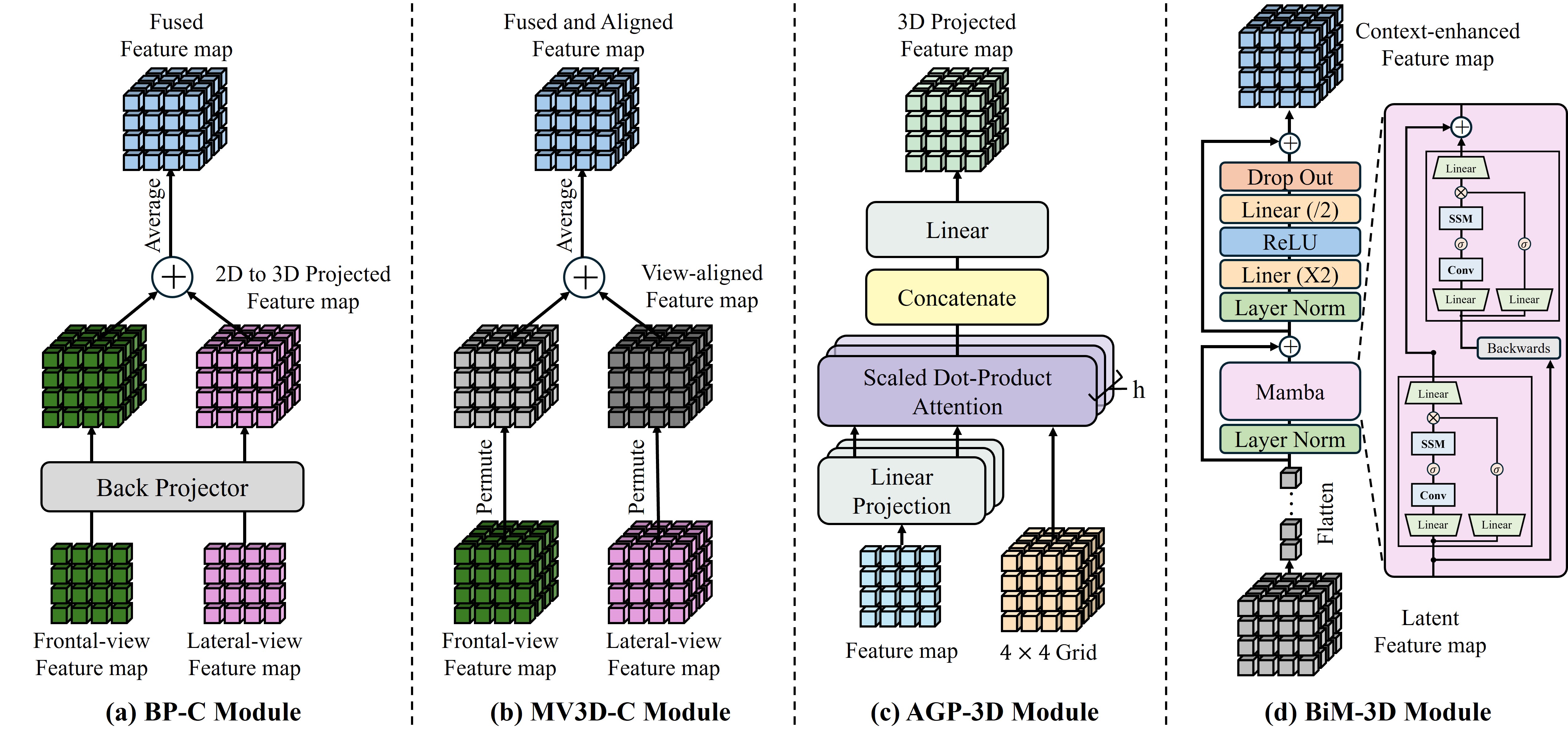}
    \caption{\textbf{Visualization of proposed modules.} (a) BP-C: Back-projects and fuses 2D features for encoder conditioning. (b) MV3D-C: Aligns and averages multi-view 3D features in the decoder. (c) AGP-3D: Maps 2D features to 3D voxels via attention-guided projection. (d) BiM-3D: Refines bottleneck features via bidirectional state space modeling.}
    \label{fig:3}
\end{figure*}

Recovering 3D volumes from 2D features suffers from severe depth ambiguity, often causing models to generate anatomically incorrect hallucinations~\cite{kak2001principles}. To mitigate this, we inject explicit geometric constraints via a differentiable back-projection layer~\cite{peng2021xraysyn} that distributes X-ray intensities along their corresponding physical ray paths, generating a coarse geometry-aligned volumetric prior $V_{\mathrm{prior}}$ (see Fig.~\ref{fig:1}). Formally, for each voxel $\mathbf{y}$, let
$\mathbf{u}=\pi_M(\mathbf{y})$ denote its corresponding detector coordinate.
The back-projection is defined as
\begin{equation}
\mathrm{BP}(I_{\mathrm{in}},M)(\mathbf{y})
=
\left\{
\begin{array}{ll}
\displaystyle
\frac{I_{\mathrm{in}}(\mathbf{u})}
     {|L_{\mathbf{u}}|\,\Delta p},
& \mathbf{u}\in\Omega_I, \\[6pt]
0,
& \mathrm{otherwise}.
\end{array}
\right.
\label{eq:bp_operator}
\end{equation}
where $\pi_M$ denotes the projection determined by the projection matrix $M$, and $\Omega_I$ is the valid detector domain. $L_{\mathbf{u}}$ denotes the set of sampled points along the ray corresponding to detector coordinate $\mathbf{u}$, and $\Delta p$ is the sampling interval. Thus, $|L_{\mathbf{u}}|\Delta p$ approximates the in-volume ray length and normalizes the intensity distributed along the ray. The resulting volume $\mathrm{BP}(I_{\mathrm{in}},M)\in\mathbb{R}^{H\times W\times D}$ provides a coarse volumetric prior aligned with the acquisition geometry.

\subsection{Geometry-aware Multi-view Conditioning}

While the initial back-projection provides a spatial guide, the network must continuously reference the original acquisition geometry to refine details. Relying on a single injection point is insufficient for recovering fine bone sutures. Therefore, we introduce a dual conditioning strategy that enforces geometric consistency at both the semantic encoding and structural decoding stages (Fig. \ref{fig:3}(a,b)). First, the \textbf{BP-C Module} operates at the encoder level by back-projecting 2D feature maps into 3D space according to the acquisition geometry. These view-specific volumes are averaged to form a global prior, which is concatenated with the main encoder features to resolve depth ambiguity early.

Complementarily, the \textbf{MV3D-C Module} refines the decoder features. It aligns high-level volumetric features from view-specific branches into a common coordinate system. The aligned features are averaged and concatenated with the main decoder, ensuring that the final reconstruction remains semantically consistent with the input projections.

\subsection{Attention-Guided Projection (AGP-3D) Module}

Conventional methods often utilize fixed linear projection or simple feature replication to lift 2D features into 3D space~\cite{henzler2018single,shen2019patient,chen2023bx2s,ge2022x,ying2019x2ct,wang2023trct,bai2024xctdiff,liu2024diffux2ct}. However, these approaches assume a uniform contribution of 2D pixels along the projection ray, ignoring the non-linear correspondences between specific image regions and varying depths.

To resolve this, we propose the AGP-3D module (Fig. \ref{fig:3}(c)), which leverages a Multi-Head Attention (MHA) mechanism~\cite{vaswani2017attention}. By treating 3D grid locations as queries and 2D features as keys, the network dynamically learns where to retrieve relevant texture information, enabling discriminative feature aggregation rather than blind projection.
Given a 2D feature map $x$, we flatten and project it into a sequence $x' \in \mathbb{R}^{B \times (HW) \times C}$. We then define a learnable query embedding $q \in \mathbb{R}^{B \times (D H' W') \times C}$ corresponding to the target 3D voxel grid. The geometry-aware 3D volume is synthesized via:
\begin{equation}
\label{eq:agp}
V_{\text{out}} = \text{Reshape}(\text{MultiHead}(q, x', x')),
\end{equation}
where $V_{\text{out}} \in \mathbb{R}^{B \times C \times D \times H' \times W'}$.

\subsection{Bidirectional Mamba (BiM-3D) Module}

Capturing global volumetric context is essential, yet standard Transformers scale quadratically ($O(N^2)$) with sequence length, while convolutions have limited receptive fields. We address this with the BiM-3D module (Fig. \ref{fig:3}(d)), which leverages a Bidirectional Selective State Space Model (Bi-SSM)~\cite{bick2024transformers} to model long-range dependencies with linear complexity ($O(N)$)~\cite{gu2024mamba}. Given a feature map $x \in \mathbb{R}^{B \times C \times D \times H \times W}$, we flatten spatial dimensions into a token sequence, enabling efficient global context modeling~\cite{ma2024u,ruan2024vm}.

The module comprises two residual stages: spatial mixing via bidirectional scanning and channel mixing via a feed-forward network (FFN). The process is formulated as:
\begin{equation}
\label{eq:bim}
x' = x + \text{Bi-SSM}(\text{LN}(x)), \quad x_{\text{out}} = x' + \text{FFN}(\text{LN}(x')),
\end{equation}
where $\text{LN}$ denotes Layer Normalization. Bi-SSM scans the sequence bidirectionally for global receptivity, and FFN refines channel information with SiLU activation~\cite{elfwing2018sigmoid}.

\subsection{Loss Function}
Our training objective comprises three terms to ensure volumetric fidelity, geometric consistency, and texture realism:
\begin{equation}
\mathcal{L} = \lambda_{\text{adv}} \mathcal{L}_{\text{adv}} + \lambda_{\text{rec}} \mathcal{L}_{\text{rec}} + \lambda_{\text{proj}} \mathcal{L}_{\text{proj}},
\end{equation}
where the balancing weights $(\lambda_{\text{adv}}, \lambda_{\text{rec}}, \lambda_{\text{proj}})$ are empirically set to $(0.1, 10, 10)$. We adopt a conditional Least Squares GAN (LSGAN)~\cite{mao2017least} with a 3D PatchDiscriminator~\cite{isola2017image} for $\mathcal{L}_{\text{adv}}$. $\mathcal{L}_{\text{rec}}$ is the $\ell_1$ voxel-wise reconstruction loss, and $\mathcal{L}_{\text{proj}}$ enforces consistency via 2D orthogonal projections~\cite{ying2019x2ct}.

%% file: experiments.tex
\section{Experiments}

\subsection{Datasets}

\begin{figure}[t]
    \centering
    \includegraphics[width=\linewidth]{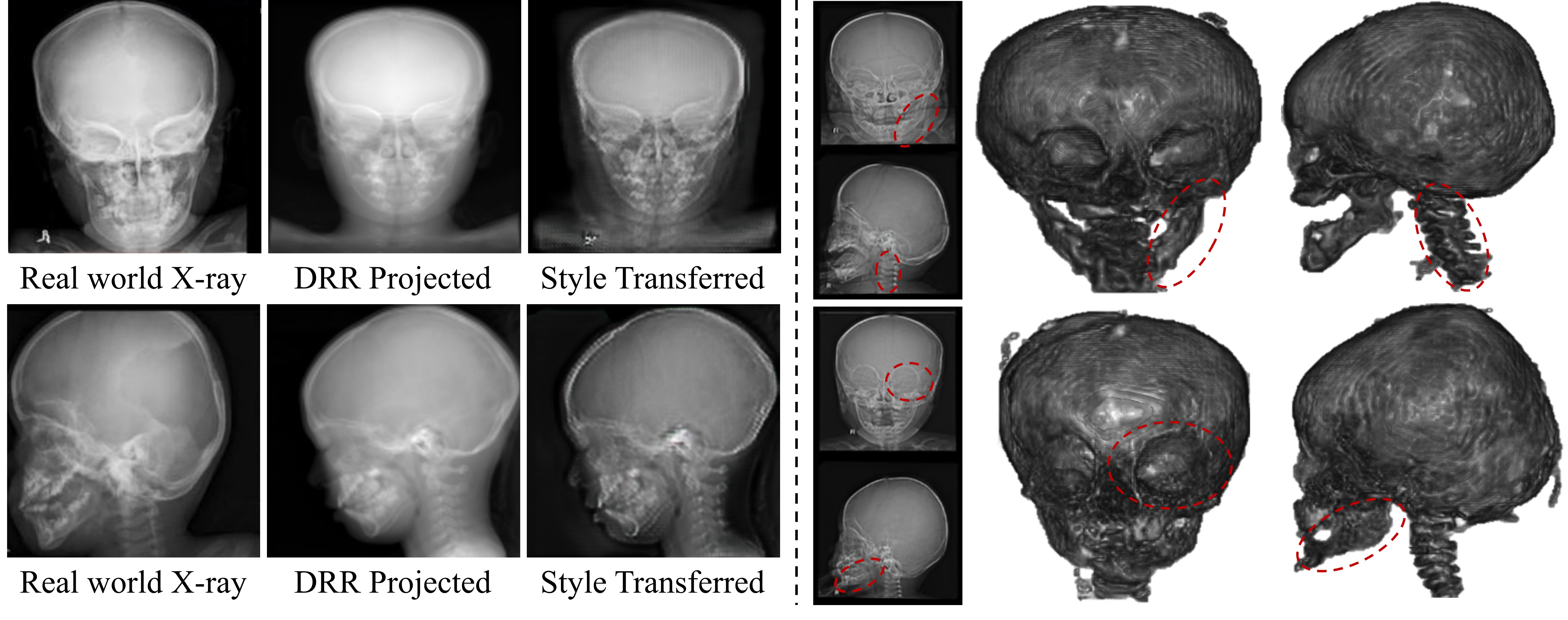}
    \caption{\textbf{Left:} Real-world, DRR~\cite{milickovic2000ct}, and style-transferred X-rays. \textbf{Right:} Qualitative reconstructions from clinical X-rays; red circles highlight corresponding anatomical regions. Paired CT ground truth was unavailable.}
    \label{fig:4}
\end{figure}

\noindent\textbf{Private PedSkull-CT.} Existing public CT datasets~\cite{armato2004lung,deng2024ctspine1k,liu2021ctpelvic} lack pediatric-specific characteristics (e.g., lower bone density, unclosed fontanelles), causing domain shift. To address this, we construct PedSkull-CT, comprising 982 CT scans from patients aged 1–24 months. Each volume is resampled to $1 \times 1 \times 1$ mm³ isotropic resolution and resized to $128^3$ voxels. Cranial bones are extracted via HU thresholding ($>$200) and min-max normalized to $[0, 1]$. Bi-planar X-ray inputs are synthesized as DRRs and translated to realistic X-ray style using CycleGAN~\cite{zhu2017unpaired} (Fig.~\ref{fig:4}). The dataset is split into 883 training and 99 test volumes.

\noindent\textbf{Public Datasets.} To demonstrate that our geometry-aware approach generalizes beyond pediatric skull imaging, we evaluate on three public benchmarks spanning diverse anatomical regions: LIDC-IDRI~\cite{armato2004lung} (thorax, 917/101 train/test, standard split~\cite{ying2019x2ct}), CTSpine1K~\cite{deng2024ctspine1k} (spine, 904/101), and CTPelvic1K~\cite{liu2021ctpelvic} (pelvis, 424/43, excluding overlaps with CTSpine1K). These datasets cover distinct skeletal structures with varying geometric complexity, enabling fair comparison with prior methods on established benchmarks.

\begin{table*}[t]
\caption{Quantitative comparison of CT reconstruction performance on public datasets.  Best results are shown in \textbf{bold}, and the second-best are \underline{underlined}.}
\label{tab:1}
\centering
\resizebox{\textwidth}{!}{%
\begin{tabular}{c|c|cccccc|c}
\hline \hline
\multirow{2}{*}{Datasets} & \multirow{2}{*}{Metrics} & ReconNet & X-CTRSNet & X2CT-GAN & TRCT-GAN & BX2S-Net & DiffuX2CT & \cellcolor{gray!10}\textbf{PSCT-Net} \\
 & &~\cite{shen2019patient} &~\cite{ge2022x} &~\cite{ying2019x2ct} &~\cite{wang2023trct} &~\cite{chen2023bx2s} &~\cite{liu2024diffux2ct} & \cellcolor{gray!10}\textbf{(Ours)} \\ 
\hline
\multirow{4}{*}{LIDC-IDRI~\cite{armato2004lung}} 
 & PSNR$\uparrow$   & 17.42 & 19.73 & 26.03 & 26.14 & 22.18 & \underline{26.35} & \cellcolor{gray!10}\textbf{27.18} \\
 & PSNR3D$\uparrow$ & 11.07 & 17.81 & \underline{21.64} & 21.43 & 20.71 & 21.15 & \cellcolor{gray!10}\textbf{22.14} \\
 & SSIM$\uparrow$   & 0.310 & 0.541 & 0.645 & 0.632 & 0.556 & \textbf{0.687} & \cellcolor{gray!10}\underline{0.671} \\
 & LPIPS$\downarrow$& 0.698 & 0.388 & \underline{0.114} & 0.120 & 0.463 & 0.122 & \cellcolor{gray!10}\textbf{0.102} \\ 
\hline
\multirow{4}{*}{CTSpine1K~\cite{deng2024ctspine1k}} 
 & PSNR$\uparrow$   & 18.01 & 18.81 & 20.82 & \underline{21.57} & 20.76 & 21.53 & \cellcolor{gray!10}\textbf{21.86} \\
 & PSNR3D$\uparrow$ & 17.57 & 18.36 & 20.23 & 20.95 & 20.25 & \underline{21.12} & \cellcolor{gray!10}\textbf{21.15} \\
 & SSIM$\uparrow$   & 0.395 & 0.394 & 0.425 & 0.456 & 0.463 & \textbf{0.592} & \cellcolor{gray!10}\underline{0.511} \\
 & LPIPS$\downarrow$& 0.7078 & 0.631 & 0.223 & 0.211 & 0.641 & \textbf{0.167} & \cellcolor{gray!10}\underline{0.199} \\ 
\hline
\multirow{4}{*}{CTPelvic1K~\cite{liu2021ctpelvic}} 
 & PSNR$\uparrow$   & 30.49 & 13.86 & \underline{31.71} & 30.97 & 30.34 & 23.91 & \cellcolor{gray!10}\textbf{33.06} \\
 & PSNR3D$\uparrow$ & 25.86 & 13.23 & \underline{28.53} & 27.80 & 27.62 & 23.27 & \cellcolor{gray!10}\textbf{29.03} \\
 & SSIM$\uparrow$   & 0.622 & 0.213 & \underline{0.753} & 0.746 & 0.700 & 0.658 & \cellcolor{gray!10}\textbf{0.786} \\
 & LPIPS$\downarrow$& 0.465 & 0.651 & \underline{0.113} & 0.115 & 0.465 & 0.169 & \cellcolor{gray!10}\textbf{0.108} \\ 
\hline \hline
\end{tabular}%
}
\end{table*}

\begin{table*}[t]
    \centering
    \begin{minipage}[t]{0.64\textwidth}
        \caption{Quantitative comparison on the private PedSkull-CT cohort.}
        \label{tab:2}
        \centering
        \small
        \resizebox{\textwidth}{!}{%
        \begin{tabular}{lcccc}
            \toprule
            \textbf{Methods} & \textbf{PSNR}$\uparrow$ & \textbf{PSNR3D}$\uparrow$ & \textbf{SSIM}$\uparrow$ & \textbf{LPIPS}$\downarrow$ \\
            \midrule
            ReconNet~\cite{shen2019patient} & 25.22 & 23.04 & 0.633 & 0.537 \\
            X-CTRSNet~\cite{ge2022x} & 24.27 & 22.72 & 0.788 & 0.498 \\
            X2CT-GAN~\cite{ying2019x2ct} & \underline{30.21} & 26.61 & \underline{0.860} & \underline{0.113} \\
            TRCT-GAN~\cite{wang2023trct} & 29.92 & 26.12 & 0.847 & 0.123 \\
            BX2S-Net~\cite{chen2023bx2s} & 29.73 & \underline{26.77} & 0.777 & 0.298 \\
            \rowcolor{gray!10} \textbf{PSCT-Net} & \textbf{31.49} & \textbf{28.03} & \textbf{0.882} & \textbf{0.100} \\
            \bottomrule
        \end{tabular}%
        }
    \end{minipage}
    \hfill 
    \begin{minipage}[t]{0.34\textwidth}
        \caption{Ablation study on the LIDC-IDRI dataset.}
        \label{tab:3}
        \centering
        \small
        \resizebox{\textwidth}{!}{%
        \begin{tabular}{cccccc}
            \toprule
            \multicolumn{4}{c}{\textbf{Modules}} & \multicolumn{2}{c}{\textbf{Metrics}} \\
            \cmidrule(lr){1-4} \cmidrule(lr){5-6}
            \textbf{BP-I} & \textbf{GO-C} & \textbf{AGP} & \textbf{BiM} & \textbf{PSNR}$\uparrow$ & \textbf{SSIM}$\uparrow$ \\
            \midrule
             & & & & 26.03 & 0.645 \\
            \checkmark & & & & 26.30 & 0.647 \\
            \checkmark & \checkmark & & & 26.40 & 0.648 \\
             & & \checkmark & & 26.92 & 0.649 \\
             & & & \checkmark & 27.07 & 0.665 \\
            \checkmark & \checkmark & \checkmark & & 26.85 & 0.651 \\
            \checkmark & \checkmark & & \checkmark & \underline{27.16} & \underline{0.669} \\
            \checkmark & \checkmark & \checkmark & \checkmark & \textbf{27.18} & \textbf{0.671} \\
            \bottomrule
        \end{tabular}%
        }
    \end{minipage}
\end{table*}

\subsection{Comparisons with SOTA methods}

\renewcommand{\thefootnote}{\fnsymbol{footnote}}

\noindent\textbf{Performance on Public Benchmarks.} As shown in Table~\ref{tab:1}, PSCT-Net achieves highly competitive performance across diverse anatomical regions. On LIDC-IDRI, our method surpasses the diffusion-based DiffuX2CT~\cite{liu2024diffux2ct}\footnote{DiffuX2CT results in Table~\ref{tab:1} are quoted from the original paper; it is excluded from Table~\ref{tab:2} due to code unavailability.} in PSNR (+0.83~dB) and LPIPS, while remaining competitive in SSIM. On CTPelvic1K, our method achieves 33.06~dB PSNR, outperforming the second-best by 1.35~dB. While iterative diffusion models may occasionally yield higher SSIM in certain structures (e.g., CTSpine1K) due to their extensive generative priors, PSCT-Net achieves comparable or superior overall fidelity in a single forward pass, maintaining crucial inference efficiency (achieving over 1.52 volumes/s; batch size 1, on an NVIDIA H100 80GB).

\noindent\textbf{Performance on PedSkull-CT Cohort.} In Table~\ref{tab:2}, PSCT-Net outperforms all baselines, improving PSNR by 1.28~dB, SSIM by 0.022 and LPIPS~\cite{zhang2018unreasonable} by 0.013 over the second-best X2CT-GAN. Since the PedSkull-CT target volumes are explicitly thresholded to isolate osseous structures (HU > 200), these improvements directly reflect superior bone reconstruction fidelity.

\noindent\textbf{Qualitative Feasibility on Real-world X-rays.} We qualitatively applied PSCT-Net to clinical X-rays unseen during reconstruction training. Fig.~\ref{fig:4} shows anatomically plausible outputs under domain shift; however, paired CT ground truth was unavailable, and this experiment therefore does not establish quantitative real-world generalization.

\noindent\textbf{Visual Comparisons.} Fig.~\ref{fig:5} compares PSCT-Net with competing methods. Our method recovers fine-grained details on LIDC-IDRI and preserves delicate structures on PedSkull-CT that are often distorted by prior models.

\begin{figure}
    \centering
    \includegraphics[width=\linewidth]{figures/Figure5.jpg}
    \caption{\textbf{Qualitative comparison of CT reconstructions} on the LIDC-IDRI~\cite{armato2004lung} (Left) and the private PedSkull-CT cohort (Right). Red dashed regions highlight representative structural differences among the methods.}
    \label{fig:5}
    \vspace{-4mm}
\end{figure}

\subsection{Ablation Studies}

Table~\ref{tab:3} summarizes the ablation results, where GO-C jointly denotes the encoder-side BP-C and decoder-side MV3D-C pathways. BiM-3D provides the largest single-component gain (+1.04~dB). The full model achieves 27.18~dB PSNR and 0.671 SSIM, exceeding BiM-3D alone by 0.11~dB and 0.006 SSIM, respectively. These results indicate that BiM-3D is the primary contributor to the observed gains, while the geometry-aware modules provide a modest complementary benefit.

%% file: conclusion.tex
\section{Conclusion}

We presented PSCT-Net, a geometry-aware framework for reconstructing 3D CT volumes from bi-planar X-rays. By integrating differentiable back-projection with attention-guided feature lifting and efficient Bidirectional Mamba modules, our method explicitly models acquisition geometry while maintaining computational feasibility for clinical use. We also curated PedSkull-CT, a private institutional pediatric skull CT cohort, to evaluate reconstruction performance in a clinically relevant pediatric cranial setting. Experiments across public benchmarks and PedSkull-CT demonstrate highly competitive performance and superior inference efficiency, suggesting that geometry-aware reconstruction offers a viable path toward low-dose pediatric imaging.

\noindent\textbf{Limitation and Future Work.} While PSCT-Net recovers global geometry and major osseous structures with high fidelity, resolving sub-millimeter details such as fine cranial sutures remains challenging due to the current $128^3$ voxel resolution constraints. Future work will scale the framework to higher resolutions (e.g., $256^3$), conduct extensive quantitative evaluations on real-world X-rays with robust domain adaptation, and validate clinical utility through reader studies on craniosynostosis diagnosis.

%% file: cas-refs.bib
@inproceedings{ying2019x2ct,
  title={X2CT-GAN: reconstructing CT from biplanar X-rays with generative adversarial networks},
  author={Ying, Xingde and Guo, Heng and Ma, Kai and Wu, Jian and Weng, Zhengxin and Zheng, Yefeng},
  booktitle={Proceedings of the IEEE/CVF conference on computer vision and pattern recognition},
  pages={10619--10628},
  year={2019}
}

@article{wang2023trct,
  title={TRCT-GAN: CT reconstruction from biplane X-rays using transformer and generative adversarial networks},
  author={Wang, Yufeng and Sun, Zhan-Li and Zeng, Zhigang and Lam, Kin-Man},
  journal={Digital Signal Processing},
  volume={140},
  pages={104123},
  year={2023},
  publisher={Elsevier}
}

@article{song2020denoising,
  title={Denoising diffusion implicit models},
  author={Song, Jiaming and Meng, Chenlin and Ermon, Stefano},
  journal={arXiv preprint arXiv:2010.02502},
  year={2020}
}

@inproceedings{liu2024diffux2ct,
  title={Diffux2ct: Diffusion learning to reconstruct CT images from biplanar x-rays},
  author={Liu, Xuhui and Qiao, Zhi and Liu, Runkun and Li, Hong and Zhang, Juan and Zhen, Xiantong and Qian, Zhen and Zhang, Baochang},
  booktitle={European conference on computer vision},
  pages={458--476},
  year={2024},
  organization={Springer}
}

@article{bai2024xctdiff,
  title={Xctdiff: Reconstruction of ct images with consistent anatomical structures from a single radiographic projection image},
  author={Bai, Qingze and Liu, Tiange and Liu, Zhi and Tong, Yubing and Torigian, Drew and Udupa, Jayaram},
  journal={arXiv preprint arXiv:2406.04679},
  year={2024}
}

@article{chen2023bx2s,
  title={BX2S-Net: Learning to reconstruct 3D spinal structures from bi-planar X-ray images},
  author={Chen, Zheye and Guo, Lijun and Zhang, Rong and Fang, Zhongding and He, Xiuchao and Wang, Jianhua},
  journal={Computers in biology and medicine},
  volume={154},
  pages={106615},
  year={2023},
  publisher={Elsevier}
}

@article{ge2022x,
  title={X-CTRSNet: 3D cervical vertebra CT reconstruction and segmentation directly from 2D X-ray images},
  author={Ge, Rongjun and He, Yuting and Xia, Cong and Xu, Chenchu and Sun, Weiya and Yang, Guanyu and Li, Junru and Wang, Zhihua and Yu, Hailing and Zhang, Daoqiang and others},
  journal={Knowledge-Based Systems},
  volume={236},
  pages={107680},
  year={2022},
  publisher={Elsevier}
}

@inproceedings{peng2021xraysyn,
  title={Xraysyn: Realistic view synthesis from a single radiograph through ct priors},
  author={Peng, Cheng and Liao, Haofu and Wong, Gina and Luo, Jiebo and Zhou, S Kevin and Chellappa, Rama},
  booktitle={Proceedings of the AAAI Conference on Artificial Intelligence},
  volume={35},
  pages={436--444},
  year={2021}
}

@article{vaswani2017attention,
  title={Attention is all you need},
  author={Vaswani, Ashish and Shazeer, Noam and Parmar, Niki and Uszkoreit, Jakob and Jones, Llion and Gomez, Aidan N and Kaiser, {\L}ukasz and Polosukhin, Illia},
  journal={Advances in neural information processing systems},
  volume={30},
  year={2017}
}

@inproceedings{mao2017least,
  title={Least squares generative adversarial networks},
  author={Mao, Xudong and Li, Qing and Xie, Haoran and Lau, Raymond YK and Wang, Zhen and Paul Smolley, Stephen},
  booktitle={Proceedings of the IEEE international conference on computer vision},
  pages={2794--2802},
  year={2017}
}

@inproceedings{isola2017image,
  title={Image-to-image translation with conditional adversarial networks},
  author={Isola, Phillip and Zhu, Jun-Yan and Zhou, Tinghui and Efros, Alexei A},
  booktitle={Proceedings of the IEEE conference on computer vision and pattern recognition},
  pages={1125--1134},
  year={2017}
}

@article{armato2004lung,
  title={Lung image database consortium: developing a resource for the medical imaging research community},
  author={Armato III, Samuel G and McLennan, Geoffrey and McNitt-Gray, Michael F and Meyer, Charles R and Yankelevitz, David and Aberle, Denise R and Henschke, Claudia I and Hoffman, Eric A and Kazerooni, Ella A and MacMahon, Heber and others},
  journal={Radiology},
  volume={232},
  number={3},
  pages={739--748},
  year={2004},
  publisher={Radiological Society of North America}
}

@article{milickovic2000ct,
  title={CT imaging based digitally reconstructed radiographs and their application inbrachytherapy},
  author={Milickovic, Natasa and Baltas, Dimos and Giannouli, S and Lahanas, M and Zamboglou, N},
  journal={Physics in Medicine \& Biology},
  volume={45},
  number={10},
  pages={2787},
  year={2000},
  publisher={IOP Publishing}
}

@inproceedings{zhu2017unpaired,
  title={Unpaired image-to-image translation using cycle-consistent adversarial networks},
  author={Zhu, Jun-Yan and Park, Taesung and Isola, Phillip and Efros, Alexei A},
  booktitle={Proceedings of the IEEE international conference on computer vision},
  pages={2223--2232},
  year={2017}
}

@inproceedings{zhang2018unreasonable,
  title={The unreasonable effectiveness of deep features as a perceptual metric},
  author={Zhang, Richard and Isola, Phillip and Efros, Alexei A and Shechtman, Eli and Wang, Oliver},
  booktitle={Proceedings of the IEEE conference on computer vision and pattern recognition},
  pages={586--595},
  year={2018}
}

@article{mettler2008effective,
  title={Effective doses in radiology and diagnostic nuclear medicine: a catalog},
  author={Mettler Jr, Fred A and Huda, Walter and Yoshizumi, Terry T and Mahesh, Mahadevappa},
  journal={Radiology},
  volume={248},
  number={1},
  pages={254--263},
  year={2008},
  publisher={Radiological Society of North America}
}

@article{brenner2007computed,
  title={Computed tomography—an increasing source of radiation exposure},
  author={Brenner, David J and Hall, Eric J},
  journal={New England journal of medicine},
  volume={357},
  number={22},
  pages={2277--2284},
  year={2007},
  publisher={Mass Medical Soc}
}

@inproceedings{henzler2018single,
  title={Single-image tomography: 3D volumes from 2D cranial x-rays},
  author={Henzler, Phlipp and Rasche, Volker and Ropinski, Timo and Ritschel, Tobias},
  booktitle={Computer Graphics Forum},
  volume={37},
  pages={377--388},
  year={2018},
  organization={Wiley Online Library}
}

@article{shen2019patient,
  title={Patient-specific reconstruction of volumetric computed tomography images from a single projection view via deep learning},
  author={Shen, Liyue and Zhao, Wei and Xing, Lei},
  journal={Nature biomedical engineering},
  volume={3},
  number={11},
  pages={880--888},
  year={2019},
  publisher={Nature Publishing Group UK London}
}

@misc{deng2024ctspine1k,
      title={CTSpine1K: A Large-Scale Dataset for Spinal Vertebrae Segmentation in Computed Tomography}, 
      author={Yang Deng and Ce Wang and Yuan Hui and Qian Li and Jun Li and Shiwei Luo and Mengke Sun and Quan Quan and Shuxin Yang and You Hao and Pengbo Liu and Honghu Xiao and Chunpeng Zhao and Xinbao Wu and S. Kevin Zhou},
      year={2024},
      eprint={2105.14711},
      archivePrefix={arXiv},
      primaryClass={eess.IV},
      url={https://arxiv.org/abs/2105.14711}, 
}

@misc{liu2021ctpelvic,
      title={Deep Learning to Segment Pelvic Bones: Large-scale CT Datasets and Baseline Models}, 
      author={Pengbo Liu and Hu Han and Yuanqi Du and Heqin Zhu and Yinhao Li and Feng Gu and Honghu Xiao and Jun Li and Chunpeng Zhao and Li Xiao and Xinbao Wu and S. Kevin Zhou},
      year={2021},
      eprint={2012.08721},
      archivePrefix={arXiv},
      primaryClass={cs.CV},
      url={https://arxiv.org/abs/2012.08721}, 
}

@article{xie2025dvg,
  title={DVG-Diffusion: Dual-View Guided Diffusion Model for CT Reconstruction from X-Rays},
  author={Xie, Xing and Liu, Jiawei and Fan, Huijie and Han, Zhi and Tang, Yandong and Qu, Liangqiong},
  journal={arXiv preprint arXiv:2503.17804},
  year={2025}
}

@inproceedings{gu2024mamba,
  title={Mamba: Linear-time sequence modeling with selective state spaces},
  author={Gu, Albert and Dao, Tri},
  booktitle={First conference on language modeling},
  year={2024}
}

@article{bick2024transformers,
  title={Transformers to SSMs: Distilling Quadratic Knowledge to Subquadratic Models},
  author={Aviv Bick and Kevin Y. Li and Eric P. Xing and J. Z. Kolter and Albert Gu},
  year={2024},
  doi={10.48550/arXiv.2408.10189},
  journal={arXiv preprint arXiv:2408.10189},
}

@article{kim2025multi,
  title={Multi-Modal and Multi-View Fusion Classifier for Craniosynostosis Diagnosis.},
  author={Kim, Dong Yeong and Kim, Joo Whan and Kim, Seung-Ki and Kim, Young-Gon},
  journal={Studies in Health Technology and Informatics},
  volume={329},
  pages={578--582},
  year={2025}
}

@article{vannier1989craniosynostosis,
  title={Craniosynostosis: diagnostic value of three-dimensional CT reconstruction.},
  author={Vannier, Michael W and Hildebolt, Charles F and Marsh, Jeffrey L and Pilgram, Thomas K and McAlister, William H and Shackelford, Gary D and Offutt, Carolyn J and Knapp, Robert H},
  journal={Radiology},
  volume={173},
  number={3},
  pages={669--673},
  year={1989}
}

@book{kak2001principles,
  title={Principles of computerized tomographic imaging},
  author={Kak, Avinash C and Slaney, Malcolm},
  year={2001},
  publisher={SIAM}
}

@article{ma2024u,
  title={U-mamba: Enhancing long-range dependency for biomedical image segmentation},
  author={Ma, Jun and Li, Feifei and Wang, Bo},
  journal={arXiv preprint arXiv:2401.04722},
  year={2024}
}

@article{ruan2024vm,
  title={Vm-unet: Vision mamba unet for medical image segmentation},
  author={Ruan, Jiacheng and Li, Jincheng and Xiang, Suncheng},
  journal={ACM Transactions on Multimedia Computing, Communications and Applications},
  year={2024},
  publisher={ACM New York, NY}
}

@misc{goske2008image,
  title={The Image Gently campaign: working together to change practice},
  author={Goske, Marilyn J and Applegate, Kimberly E and Boylan, Jennifer and Butler, Priscilla F and Callahan, Michael J and Coley, Brian D and Farley, Shawn and Frush, Donald P and Hernanz-Schulman, Marta and Jaramillo, Diego and others},
  journal={AJR. American journal of roentgenology},
  volume={190},
  number={2},
  pages={273--274},
  year={2008}
}

@article{elfwing2018sigmoid,
  title={Sigmoid-weighted linear units for neural network function approximation in reinforcement learning},
  author={Elfwing, Stefan and Uchibe, Eiji and Doya, Kenji},
  journal={Neural networks},
  volume={107},
  pages={3--11},
  year={2018},
  publisher={Elsevier}
}
